\documentclass{svproc}

\usepackage{url}

\usepackage{lipsum}
\usepackage{booktabs}
\usepackage{graphicx}
\usepackage{hyperref}
\usepackage{svg}

\newcommand{\gd}{G\`aidhlig}

\begin{document}
\mainmatter               

\title{A Modular Part-of-Speech Tagger for Scottish Gaelic using spaCy}

\titlerunning{A Modular \gd\ PoS Tagger}

\author{P\'eter Stef\'an \and Peter J\@. Barclay \and Alistair Lawson}

\authorrunning{P\'eter Stef\'an et al.} 

\tocauthor{P\'eter Stef\'an, Peter J\@. Barclay, Alistair Lawson}

\institute{School of Computing, Engineering, \& the Built Environment, \\ Edinburgh Napier University,\\ 
10 Colinton Rd, Edinburgh, Scotland, UK\\
}

\maketitle               

\begin{abstract}

Part-of-speech tagging for low-resource languages remains challenging due to limited annotated data, especially for linguistically complex languages. \gd\ (Scottish Gaelic) is a morphologically rich and endangered language with limited digital resources, making it suitable for examining a lightweight language processing approach.
This paper describes using the modular spaCy Natural Language Processing framework to build part-of-speech taggers for \gd\ using the Annotated Reference Corpus of Scottish Gaelic. We train two models with minimal pre-processing and configuration: one using a fine-grained tagset and another using a reduced coarse-grained tagset. Both models are trained without external embeddings or pre-trained language models, using only supervised learning from the available corpus.
The fine-grained model achieves 88.6\% tagging accuracy, while the coarse-grained model achieves 93.7\%. The results 
are comparable to those of the two previously published \gd\ taggers, indicating that simple, off-the-shelf language processing pipelines can demonstrate
good performance in low-resource and morphologically complex linguistic settings.

\keywords{low-resource languages; part-of-speech tagging; \gd; spaCy}
\end{abstract}

\section{Introduction}
\label{sec:introduction}
Recent advances in Natural Language Processing (NLP), particularly the rise of large language models (LLMs), have demonstrated a strong dependence on large amounts of training data as well as requiring expensive computational resources. A correlation can be found between larger training corpora and the effectiveness of the models \cite{kaplan2020scalinglawsneurallanguage}. However, this data-driven approach poses an issue for low-resource languages, where  annotated corpora are limited in quantity and availability.

\gd\ (Scottish Gaelic)  is such a low-resource language. Despite 
research interest and ongoing
language revitalisation efforts, existing datasets and NLP tools for \gd\ remain limited. Available annotated corpora, such as the Annotated Reference Corpus of Scottish Gaelic (ARCOSG), are relatively smaller than those for high-resource languages. This limits the development of robust statistical or neural language models.

In addition, \gd\ presents further challenges owing to its rich morphological structure. Grammatical information is often represented by the use of inflections and word form mutations, leading to a large variety of surface forms for a given lemma \cite{tsarfaty2013ParsingMorphologicallyRich}. Part-of-speech (PoS) tagging and other NLP tasks are particularly difficult with morphologically complex languages, especially in a low-resource setting \cite{kardava2025morphology}.
As a result, low-resource NLP systems often focus on approaches leveraging detailed linguistic knowledge, using complex architectures and specialised feature engineering. 

This paper investigates whether a lightweight, general-purpose NLP framework can facilitate the construction of PoS taggers for \gd\ and low-resource languages in general. Specifically, we explore the use of the spaCy library \cite{honnibal2020spacy} to train two tagger models directly on the ARCOSG corpus with minimal configuration or preprocessing employed. Two levels of tag granularity are tested, aiming to assess the trade-off between simplicity and performance in a low-resource morphologically complex setting. 

The main contribution of this study is a reproducible lightweight \gd\ PoS tagging architecture, using off-the-shelf NLP tools and compatible with the spaCy software ecosystem, providing a competitive baseline for future research in \gd\ and other low-resource NLP settings. 
In contrast to the dozens of varied PoS tagging systems available for widely used languages such as English (see \cite{chiche2022part} for a survey), to our knowledge only two taggers have been previously described for \gd\ (see Section~\ref{sec:related}).
We now provide now a third tagger, the first to use spaCy, making our code publicly available.

\section{Background}
Natural Language Processing is a very important part of human computer interactions. NLP techniques are used in everyday situations, such as spell checkers and speech recognition tools. The study of n-grams and statistical language modelling represented a major advancement in NLP: text corpora can be used to observe patterns and establish word sequence probabilities, which then can be used to model the internal logic of a given language~\cite{chelba2010statisticalLanguageModeling}.

\subsection{\gd}
\gd\ is an Insular Celtic language indigenous to the British Isles, primarily spoken in the Scottish Highlands and Western Isles. Despite its 
widespread former use, and its great historical and cultural significance, the number of speakers has declined in recent centuries, 
owing
to a combination of historical suppression and broader political and social processes that have favoured the widespread adoption of English, resulting in its classification as an endangered language \cite{kandler2010language}. Many revitalisation efforts have been made in response, including the development of digital and NLP-based tools in recent years. These could prove crucial in supporting language learning, improving accessibility and representation in online spaces, and importantly helping the digitisation of \gd~\cite{lamb_developing_2016}. A brief overview of the history and current status of \gd\ can be found in a recent study by Barclay \cite{peter_j_barclay_rule-based_2026}.

\gd\ presents several key challenges in the development of NLP tools, typical of many low-resource languages. The existing corpora are very limited in size and availability. One primary and invaluable resource is the Annotated Reference Corpus of Scottish Gaelic (ARCOSG), described in~\cite{Lamb2020ARCOSG}. In contrast to high resource languages, \gd\ has a much smaller digital presence overall, making the development of robust NLP tools significantly more difficult~\cite{kornai2012language}.

In addition, \gd\ is a morphologically complex language. Grammatically important information is encoded into word inflections, mutational processes, and syntactic constructions \cite{lamb2024ComprehensiveScottishGrammar,ross2016orthography}. NLP tasks are more difficult in such languages due to the added complexity of finding the lemma of words, which is often an important first processing step \cite{weller2024analyzing}. Additionally, morphologically rich languages have a much higher variety in their surface word forms, meaning that a single lexical `word' will have many variant forms occurring in texts \cite{tsarfaty2013ParsingMorphologicallyRich}. The increased variety will make each individual form less likely to occur in any given training corpus, resulting in a higher number of out of vocabulary (OOV) words --
words that appear in a given text that have not been encountered before.
Thus, a language model might have more difficulty generalising effectively \cite{kardava2025morphology}. This problem is amplified in low-resource contexts owing to the already limited occurrences of these variant forms in the more restricted annotated corpora \cite{kardava2025morphology,magueresse2020lowresourcelanguagesreviewpast}.

\subsection{Related work}
\label{sec:related}
Despite these challenges, in recent years efforts have been made to develop computational resources for \gd. These include the creation of a dependency treebank \cite{batchelor_universal_2019}, which focused on capturing the morphological complexity and key grammatical features of \gd. Projects on PoS taggers, linguistic analysis, and speech recognition applications have contributed to expanding the range of digital tools available for the language \cite{lamb_evaluating_2016,boizou2020online,evans2022developing,klejch2025asrgaelic}.
However, many existing low-resource NLP systems remain highly specialised, typically designed for a single task. They rely on custom architectures with rule-based components or language-specific feature-engineering to achieve high results. 

The first tagger available for \gd, described by Lamb \textit{et al}.~\cite{lamb_developing_2014}, uses an ensemble-based Brill model,
incorporating both stochastic and rule-based methods, organised around a \textit{backoff} mechanism. Following statistical
tagging, corrective rules are applied to increase the accuracy of the final tags.
The only other prior tagger for \gd, developed by Boizou \textit{et al}.~\cite{boizou2020online} as part of the Gaelic Linguistic Analyser (GLA) project,
is based on the scikit-learn toolkit and uses additional training data in addition to ARCOSG. This tagger uses additional features to capture grammatical structure,
including prefixes and suffixes, sentence position and collocated word forms.
Although both systems are effective, approaches requiring explicit linguistic expertise and language-specific feature-engineering can increase system complexity and require greater development effort while reducing adaptability.

These issues reflect a common pattern in low-resource NLP, where the lack of large annotated corpora leads to reliance on heavily engineered pipelines, transfer learning and multilingual pre-training approaches \cite{hedderich-etal-2021-survey}. Such approaches can introduce considerable complexity and dependence on external resources, and may introduce a bias towards structures appearing any high-resource language materials used. In the case of \gd, this can be detrimental due to structural and morphological differences between the languages. For example, when using transfer learning, the characteristic Verb-Subject-Object (VSO) word order of \gd\ may not be adequately reflected in NLP resources developed primarily for SVO (Subject-Verb-Object) languages such as English.

Taking a new approach, this study investigates using a modern, lightweight, general-purpose NLP pipeline using spaCy, trained directly on the ARCOSG corpus with minimal pre-processing and configuration. This allows us to investigate how off-the-shelf NLP tools can be adopted for \gd\ PoS tagging. Our study establishes a baseline for comparison with more specialised systems, with the main advantages being the ease of implementation and reduced system complexity. These characteristics make the approach potentially transferable across other low-resource languages.

\section{Aim}
\label{sec:aims}
Despite recent progress, \gd\ NLP remains relatively underdeveloped and is constrained by the limited annotated data and tools available. This paper explores the effectiveness of \gd\ PoS tagging in a low-resource setting, focusing on tools previously unexplored in \gd\ research. We propose the use of the spaCy architecture for efficient model development, with minimal preprocessing, and limited hyperparameter tuning. The aim is to investigate whether such a lightweight NLP architecture can provide a useful baseline for future development in \gd\ language processing.

\section{Methodology}
\label{sec:method}
Existing work on \gd\ PoS tagging has focused on statistical taggers, developing specialised models with curated resources~\cite{lamb_developing_2016,boizou2020online}. While these contributions have significantly advanced the computational resources available for \gd, they may be characterised as custom-built systems.
Here, by contrast, we evaluate a general-purpose PoS tagging framework using off-the shelf components.

Leveraging the modern NLP toolkit spaCy \cite{honnibal2020spacy} remains unexplored for \gd. This leaves open the question of whether lightweight, general-purpose pipelines could be used as a baseline system for \gd\ PoS tagging and other downstream NLP tasks.
 
To address this gap, this study presents the first evaluation of a spaCy-based part-of-speech tagging pipeline for \gd. We propose two models trained on the ARCOSG corpus with different levels of PoS tag granularity for wider evaluation. Both are implemented with very limited tuning, aiming to assess the baseline performance of lightweight models for low-resource PoS tagging. The results are reproducible, providing a useful basis for future \gd\  NLP research.

\subsection{spaCy and the Pipeline Architecture}

\begin{figure}[htbp]
    \centering
    \includegraphics[width=\textwidth]{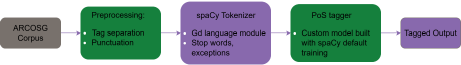}
    \caption{Overview of the \gd\ PoS tagging pipeline.}
    \label{fig:spacy_pipeline}
\end{figure}

As the core NLP framework, we used spaCy \cite{honnibal2020spacy}, an open-source library for industrial-strength NLP. The library provides a modular pipeline architecture that supports a tokeniser, parser, tagger and other components for further downstream NLP tasks. The creators of spaCy also provide a number of pre-trained models and pipelines, mostly for high-resource languages. Here, spaCy was selected owing to its modular architecture and existing NLP utilities, which reduced and simplified implementation, while allowing for language-specific customisation. Rather than requiring individual implementations of tokenisation, feature extraction, training, inference, and evaluation, these are provided through a cohesive framework. The proposed pipeline is illustrated in Figure~\ref{fig:spacy_pipeline}. Consequently, the implementation mainly focused on configuring an existing framework, assembling pipeline components and preparing language specific resources, rather than developing task-specific algorithms from scratch. This lowers the technical barrier for future \gd\ NLP research and simplifying the development of additional language technologies.

This approach is highly adaptable, so future work can build upon and extend the proposed system requiring only the development of additional components, such as a lemmatiser, dependency parser or other language tools. This enables future \gd\ NLP research to build incrementally upon the proposed system, instead of developing independent task-specific implementations, avoiding the need to redesign the overall architecture for each new project.

Most importantly, the spaCy framework provides a \textit{lang} module containing 
language-specific information for the tokeniser. Thanks to the recent work of researchers at Cornell 
College~\cite{gaelic_lemmatizer_cornell}, the \gd\ module now includes a list of stop-words and tokeniser exceptions. The tokeniser also uses a rule set shared across most languages as a default in spaCy\footnote{\gd\ language folder for spaCy: \url{https://github.com/explosion/spaCy/tree/master/spacy/lang/gd}}. 
In addition to these built-in resources, the proposed pipeline used some auxiliary rules for punctuation handling required to ensure compatibility with the annotation conventions used in ARCOSG. 

The pipeline was designed to be lightweight to reflect the low-resource conditions of \gd. Computationally expensive architectures, such as transformer-based models were excluded from consideration, since they typically require much larger training corpora for effective performance. In contrast, this study focuses on a compact supervised pipeline that can be trained from the annotated ARCOSG corpus, with minimal pre-processing and limited computational overhead, while still capturing the rich morphology of \gd.

The proposed pipeline comprises three main stages: tokenisation of the raw text, contextual feature extraction using Tok2Vec, and PoS prediction using spaCy's tagger component. Each stage is described below.

Tok2Vec\footnote{Tok2Vec documentation is available on the spaCy website: \url{https://spacy.io/api/tok2vec}} is the first trainable component of the pipeline. It converts each word from the raw input text into a numerical representation which captures both the word itself and its surrounding context. These representation are based on features such as word shape 
and the normalised form (a standardised form of the word used by spaCy, similar to the lemma). This helps the model encode important morphological information. The representations are then used as input by the tagger, which predicts the most likely PoS label for each token.

This architecture of spaCy is well suited to the morphological complexity of \gd. Grammatical information is learned from subword structures, such as affixes and other orthographic features, during the training process. This allows the model to generalise to previously unseen tokens -- with low-resource languages, it is common to encounter words and 
surface forms that did not appear in the training data
\cite{magueresse2020lowresourcelanguagesreviewpast}. PoS predictions are also heavily affected by syntactic context, making the contextual token representations from the Tok2Vec component especially important in such low-resource settings, where the corpus may otherwise contain insufficient examples~\cite{hedderich-etal-2021-survey}.

The context encoding component only considers local contextual information. This design prioritises vocabulary interactions nearer to the token, where PoS tag distinctions are resolved through the immediate context and the morphological cues discussed above. As a result, the model learns through the relationship between context and morphology, rather than the meaning of the entire sentence or text. This also reduces model complexity, making the approach more suitable for the limited amount of training data available.

We use no pre-trained vectors or external embeddings; the model must rely purely on gold-standard supervised labels from the annotated corpus.

Given the limited training data, this lightweight pipeline architecture focuses on local contextual and morphological representations during training. This helps efficient learning in a low-resource context, while maintaining attention on orthographic and syntactic indicators that are relevant for PoS tagging \cite{jurafsky_martin_2026}.

\newpage

\subsection{Data}
The entire ARCOSG corpus~\cite{Lamb2020ARCOSG} was used for training and evaluation in this study. An 80-10-10 split was applied for training, validation and testing. 
Following the approach of prior studies~\cite{lamb_developing_2016} and~\cite{boizou2020online}, we use both a fine-grained (242 or 246 items) and a coarse-grained (40 or 41 items) tag set for training and evaluation, making the comparison between previous work and our proposed models simpler. The coarse-grained tags simplify some linguistic distinctions by grouping together a set of similar fine-grained tags. We opted for using all available data to maximise the training and test sets; since no pre-training is done, the model is solely trained through supervised learning on the annotated corpus.

\subsection{Preprocessing}
\label{sec:preprocessing}
Consistent with the aim of the study, pre-processing was intentionally kept minimal in order to evaluate the effectiveness of a lightweight NLP pipeline using a spaCy-based structure in a low-resource context. No stemming or linguistic simplification was applied, to retain as much morphological information as possible, which is essential for morphologically complex PoS tagging \cite{tsarfaty2013ParsingMorphologicallyRich,kardava2025morphology}. 

The preprocessing focused on tokenisation consistency and compatibility with the ARCOSG annotation system, while preserving orthographic and morphological structure. This minimal pre-processing was intended to keep the experiments reproducible and reduce over-dependence on language-specific modelling.

\section{Results}

Both tagging models were evaluated using a 10\% randomly selected sample from  ARCOSG (the unseen test data set). 
The main metric used to measure performance is tagging accuracy, meaning the percentage of PoS tags that were correctly predicted by the model. The model using the larger ARCOSG tagset achieved an accuracy of 88.60\%. The other one, trained on the reduced set of tags, performed with 93.7\% accuracy. 
These results are comparable with the accuracy of previously reported \gd\ PoS taggers.
Table \ref{table:1} presents the model architecture of both models and their performance. 

\label{sec:results}
\begin{table}[ht]
\centering
\begin{tabular*}{\textwidth}{@{\extracolsep{\fill}}lrr}
\hline
& Full Tagset Model & Reduced Tagset Model \\
\hline
Tagset Type & Fine-grained PoS tags &  Coarse PoS tags\\
Tagset Size & 215 & 40 \\
Training Tokens & 70,890 & 69,522 \\
Vocabulary Size & 7160 & 7114 \\
\textbf{POS Tag Accuracy} & \textbf{88.60\%} & \textbf{93.70\%} \\
\hline
\end{tabular*}
\vspace{0.5em}
\caption{Performance and training set comparison between the full and reduced ARCOSG tagset models.}
\label{table:1}
\end{table}

\newpage
Two models were evaluated, using the coarse- and the fine-grained tagset respectively. The full tagset has 215 tags, the reduced set 40 coarse-grained tags. The training corpus contains 70,890 total words, with a vocabulary of 7160 unique entries; this is slightly smaller for the reduced set. The spaCy architecture even gives a warning when the training set is validated, indicating that the current number of examples may not be sufficient for proper training.

Some minor tokenisation inconsistencies remained, as $\sim$0.4\% of the tokens are labelled ``misaligned''. However, this was expected, as only the stop words and basic tokenisation exceptions were applied from the \gd\ language defaults in the spaCy framework. 

\section{Discussion}
\label{sec:discussion}

Both models show encouraging performance, with the full fine-grained model achieving results comparable to earlier specialised systems of Lamb \textit{et al}.~\cite{lamb_evaluating_2016} and Boizou \textit{et al}.~\cite{boizou2020online}. The results in Table \ref{table:2} suggest that a lightweight spaCy-based pipeline can achieve competitive performance in low-resource settings for morphologically complex languages with minimal language-specific engineering. 

The difference in performance between the full and reduced tagsets highlights the morphological complexity of \gd, as the full tagset requires the model to distinguish between much finer grammatical categories. The reduced set collapses linguistically subtle and context dependent distinctions making the classification task substantially simpler. In the low-resource setting used in these experiments, many of the 215 fine-grained tags appear infrequently: this further increases the difficulty of this task. Without pre-training or external word embeddings, the model must solely rely on corpus examples, which limits its ability to learn the more complex PoS tags.

\begin{table}[htbp]
\centering
\begin{tabular}{lll r}
\toprule
Study & Tagset Type & Tagset Size & Accuracy \\
\midrule
Danso \& Lamb (2014)  & Fine-grained   & 242 tags (PAROLE tagset) & 76.6 \\
Lamb \textit{et al.} (2016) & Fine-grained & 246 tags & 84\% \\
Lamb \textit{et al.} (2016)& Coarse-grained  & 41 tags &  92\% \\
Boizou \& Lamb (2020) & Fine-grained   & 246 tags           & 90.7\% \\
Boizou \& Lamb (2020) & Coarse-grained & 41 tags            & 94.7\%  \\
Our model         & Fine-grained   & 246 tags           & \textbf{88.6\%} \\ 
Our model          & Coarse-grained & 40 tags           & \textbf{93.7\%} \\
\bottomrule
\end{tabular}
\vspace{0.5em}
\caption{Performance comparison of previous and proposed models. Lamb \textit{et al.}\cite{lamb_evaluating_2016} provided additional evaluation on the same model developed by Danso \& Lamb\cite{DansoLamb2014GaelicPOS}.}
\label{table:2}
\end{table}

These findings are significant given the simplicity of the implementation and the model architecture. The application of the spaCy framework appears to perform effectively with minimal configuration. The models are able to learn morphological patterns, suggesting that the Tok2Vec architecture can generalise reasonably well in low-resource settings through limited local context, subword and orthographic information.

The design decision to preserve morphological information during preprocessing is also supported by these results. As noted, no stemming or lemmatisation was applied, allowing the model to learn directly from the various surface forms. The relatively strong performance,  even with the full tagset, could suggest that preserving the surface features is just as beneficial as more customised approaches commonly used in low-resource tagging tasks.

These experiments were designed to be reproducible, as the default settings were applied in most cases.
A future study could investigate customisation of the implementation, to 
investigate the best pipeline structure for \gd\ using spaCy.

Our results suggest that a lightweight, general-purpose NLP pipeline can provide a practical and reproducible baseline for \gd\ PoS tagging. The use of spaCy significantly reduced the implementation difficulty of the proposed pipelines. The use of two sets of 
tags with differing
granularity highlights the importance of tokenisation quality when addressing the complex grammatical structures in \gd.

For \gd\ and other languages with complex morphology and limited resources, these results are promising for future experiments with similar pipelines using the spaCy architecture.

\section{Further Work}

As this research is ongoing, we hope to implement similar pipelines using spaCy with more sophisticated architectures. Ideas include introducing pre-training via multilingual models such as XLM-RoBERTA\footnote{\url{https://huggingface.co/docs/transformers/model_doc/xlm-roberta}}. This would allow evaluation of whether multilingual models provide a better starting state. Some models, such as XLM-RoBERTA, also include \gd\ texts in their training data, perhaps making them a better choice. On the other hand, this approach could also introduce other limitations, such as a bias towards the grammatical structure of high-resource languages in the training set leading to negative transfer \cite{wang-etal-2020-negative-transfer}. Future work could also investigate whether using cross-lingual transfer would provide any improvements for \gd\ models over the lightweight architecture proposed here. 

Using additional data sources for unsupervised learning could be employed to increase the learning potential of future models. Given the limited available annotated corpora, this approach provides a more direct way of addressing the issue. Unlabelled \gd\ texts could be used for self-supervised pre-training, providing the models with further examples to learn lexical and morphological patterns before tuning them with the annotated data. Data augmentation techniques, such as synthetic sentence generation, paraphrasing, morphology-aware transformations, and synonym replacement, could also increase the diversity in the training set. These approaches, in the case of a language with complex morphology, may lead to improved generalisation \cite{sahin2018dataAugmentation}.

Additional analysis of the model predictions could pinpoint which tags are the most frequently misclassified and which grammatical distinctions are the most challenging. The significant performance difference shown between the full and reduced tagsets suggests that most errors may occur because of fine-grained highly specialised grammatical categories. Detailed error analysis could provide valuable insight into whether errors primarily arise between related fine-grained categories or broader syntactic ones.  Understanding the confusion patterns of the models could guide future improvements in annotation schemes, model architectures, and preprocessing strategies \cite{liu2007heuristic}. 

\section{Conclusion}
\label{sec:conclusion}
This paper presents our evaluation of the effectiveness of lightweight PoS tagging pipelines based on the spaCy architecture for \gd\ using the ARCOSG corpus. The goal was to assess whether a model with minimal preprocessing and configuration could perform as well as the more heavily engineered approaches common in low-resource NLP.

The results show that both models achieved relatively strong performance. The first model using the fine-grained 215 tagset achieved 88.6\%, while the reduced coarse-grained 40-tag model had an accuracy of 93.7\%. These indicate that the proposed pipeline using the spaCy presets in most cases is still capable of learning complex morphological and contextual representations for \gd\ tags, despite the limited data available for training. The difference in accuracy between the two models is due to the variation in tag granularity, highlighting the need for additional information to address subtle grammatical distinctions. 

Overall, this work contributes a reproducible baseline for \gd\ PoS tagging, using the spaCy architecture with minimal configuration needed, which is a new approach for the language. The results are promising as they are comparable 
in accuracy
to previous work reported in the literature, which employed more complex architectures and specific configuration for \gd\ grammatical structures. As new linguistic resources become available for \gd, baseline systems such as ours can provide a foundation for future research into more advanced NLP solutions and downstream language technologies to support the digital presence of this living, culturally important but under-resourced language.

\section*{Code Availability Statement} 
All code from this research is made publicly available in the GitHub repository at: \url{https://github.com/Speter011/A-Modular-Language-Model-for-Scottish-Gaelic}.

%
%

\newpage

\bibliographystyle{plain}
{\small
\bibliography{GaelicRefs}}

\end{document}